\pdfoutput=1
\documentclass{article}

\usepackage[preprint]{neurips_2025}

\usepackage[most]{tcolorbox}  
\usepackage{lipsum}    
\usepackage{subcaption}
\usepackage{array}
\usepackage[utf8]{inputenc} 
\usepackage[T1]{fontenc}   
\usepackage{hyperref}      
\usepackage{url}           
\usepackage{booktabs}     
\usepackage{amsfonts}       
\usepackage{nicefrac}       
\usepackage{microtype}      
\usepackage{xcolor}        
\usepackage{booktabs}      
\usepackage{amsfonts}    
\usepackage{nicefrac}      
\usepackage{microtype}    
\usepackage{xcolor}        
\usepackage{times}
\usepackage{latexsym}
\usepackage{graphicx}
\usepackage{amsmath}
\usepackage{booktabs}
\usepackage{graphicx}
\usepackage{float} 
\usepackage[normalem]{ulem}
\usepackage{xspace}

\makeatletter
\newcommand\thanksnomark[1]{
  \begingroup
  \renewcommand\@makefnmark{}
  \thanks{#1}
  \endgroup
}
\makeatother

\useunder{\uline}{\ul}{}

\usepackage{pifont}
\usepackage{wrapfig}
\definecolor{inkblue}{RGB}{40,120,181} 
\definecolor{inkred}{RGB}{200,36,35} 
\newcommand{\name}{\texttt{MedOrch}\xspace}

\title{\name: Medical Diagnosis with Tool-Augmented Reasoning Agents for Flexible Extensibility}

\author{
Yexiao He$^{1}$ \quad Ang Li$^{1}$  \quad Boyi Liu$^{2}$ \quad
Zhewei Yao$^{2,*}$\thanksnomark{Project lead.} \quad Yuxiong He$^{2}$ \\
$^1$University of Maryland \\
$^2$Snowflake \\
\texttt{\{yexiaohe,angliece\}@umd.edu} \\
\texttt{\{boyi.liu,zhewei.yao,yuxiong.he\}@snowflake.com}
}

\begin{document}

\maketitle

\begin{abstract}
Healthcare decision-making represents one of the most challenging domains for Artificial Intelligence (AI), requiring the integration of diverse knowledge sources, complex reasoning, and various external analytical tools. Current AI systems often rely on either task-specific models, which offer limited adaptability, or general language models without grounding with specialized external knowledge and tools. We introduce \name, a novel framework that orchestrates multiple specialized tools and reasoning agents to provide comprehensive medical decision support. \name employs a modular, agent-based architecture that facilitates the flexible integration of domain-specific tools without altering the core system. Furthermore, it ensures transparent and traceable reasoning processes, enabling clinicians to meticulously verify each intermediate step underlying the system’s recommendations. We evaluate \name across three distinct medical applications: Alzheimer's disease diagnosis, chest X-ray interpretation, and medical visual question answering, using authentic clinical datasets. The results demonstrate \name's competitive performance across these diverse medical tasks. Notably, in Alzheimer's disease diagnosis, \name achieves an accuracy of 93.26\%, surpassing the state-of-the-art baseline by over four percentage points. For predicting Alzheimer's disease progression, it attains a 50.35\% accuracy, marking a significant improvement. In chest X-ray analysis, \name exhibits superior performance with a Macro AUC of 61.2\% and a Macro F1-score of 25.5\%. Moreover, in complex multimodal visual question answering (Image+Table), \name achieves an accuracy of 54.47\%. These findings underscore \name's potential to advance healthcare AI by enabling reasoning-driven tool utilization for multimodal medical data processing and supporting intricate cognitive tasks in clinical decision-making.
\end{abstract}
\begin{figure*}[htbp]
  \centering

  \begin{subfigure}[b]{0.48\textwidth}
    \centering
    \includegraphics[width=\textwidth]{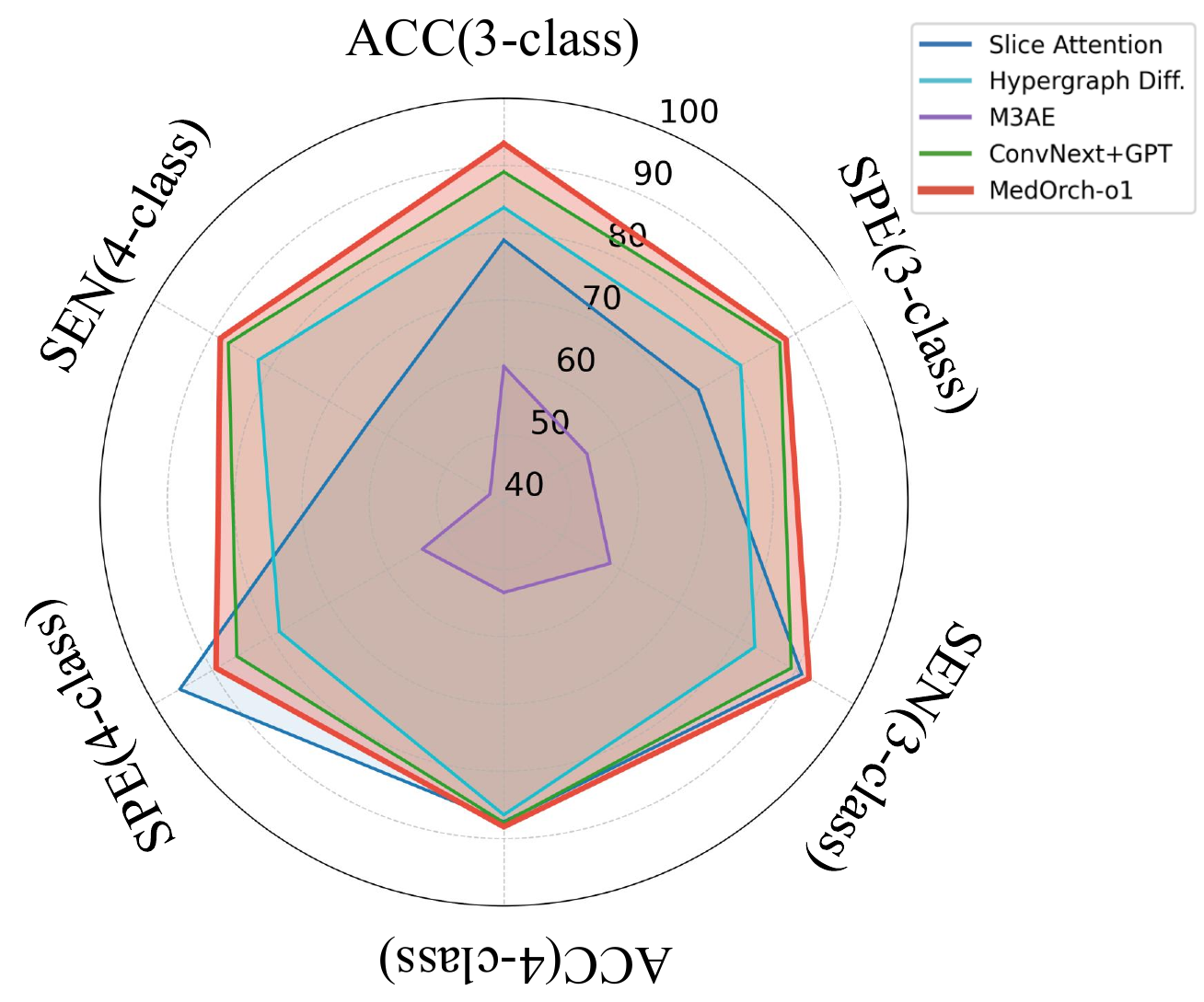}
    \caption{AD Diagnosis}
    \label{fig:radar-table1}
  \end{subfigure}
  \hfill
  \begin{subfigure}[b]{0.48\textwidth}
    \centering
    \includegraphics[width=\textwidth]{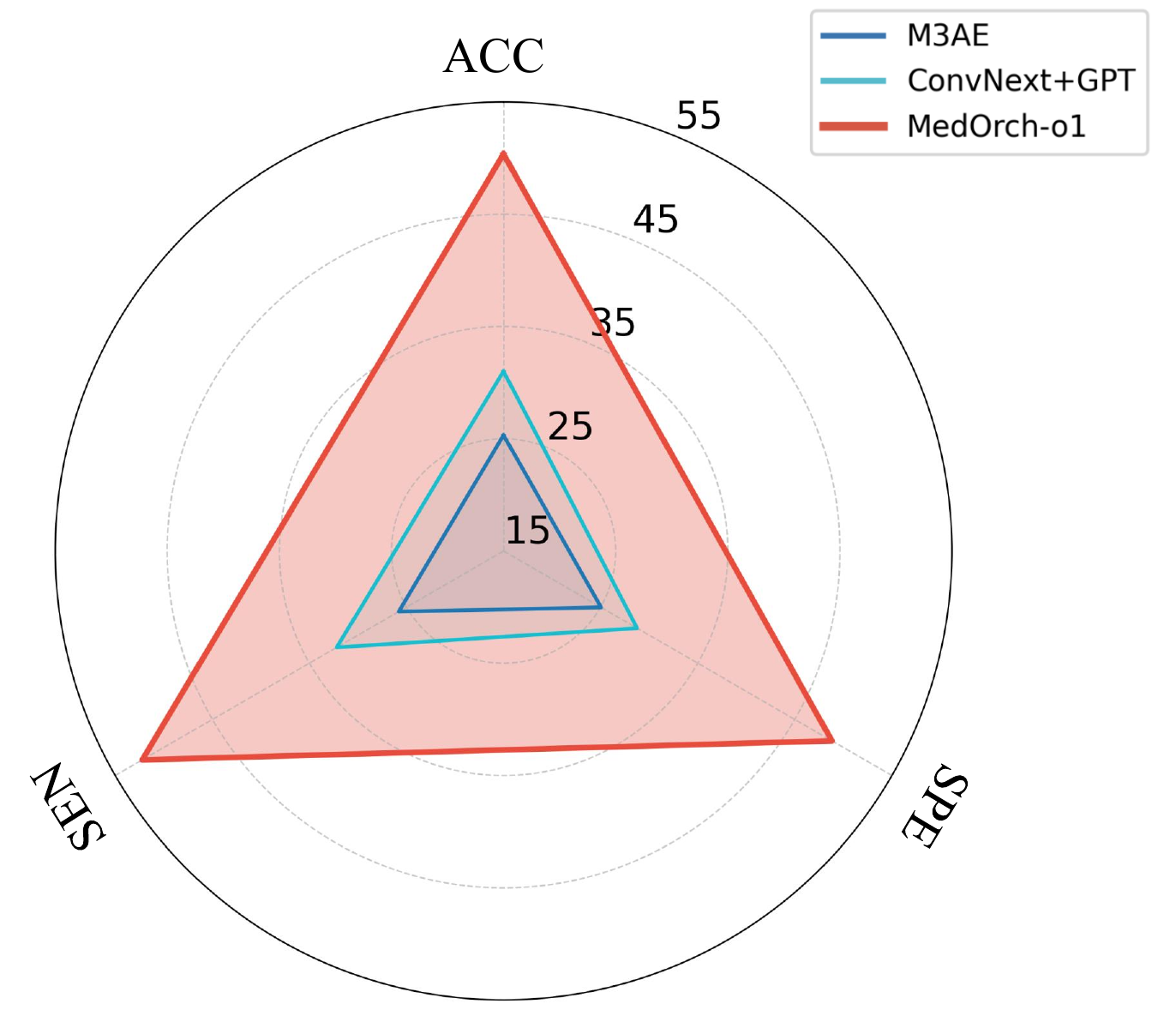}
    \caption{AD Prediction}
    \label{fig:radar-table2}
  \end{subfigure}

  \vspace{0.5em}

  \begin{subfigure}[b]{0.48\textwidth}
    \centering
    \includegraphics[width=\textwidth]{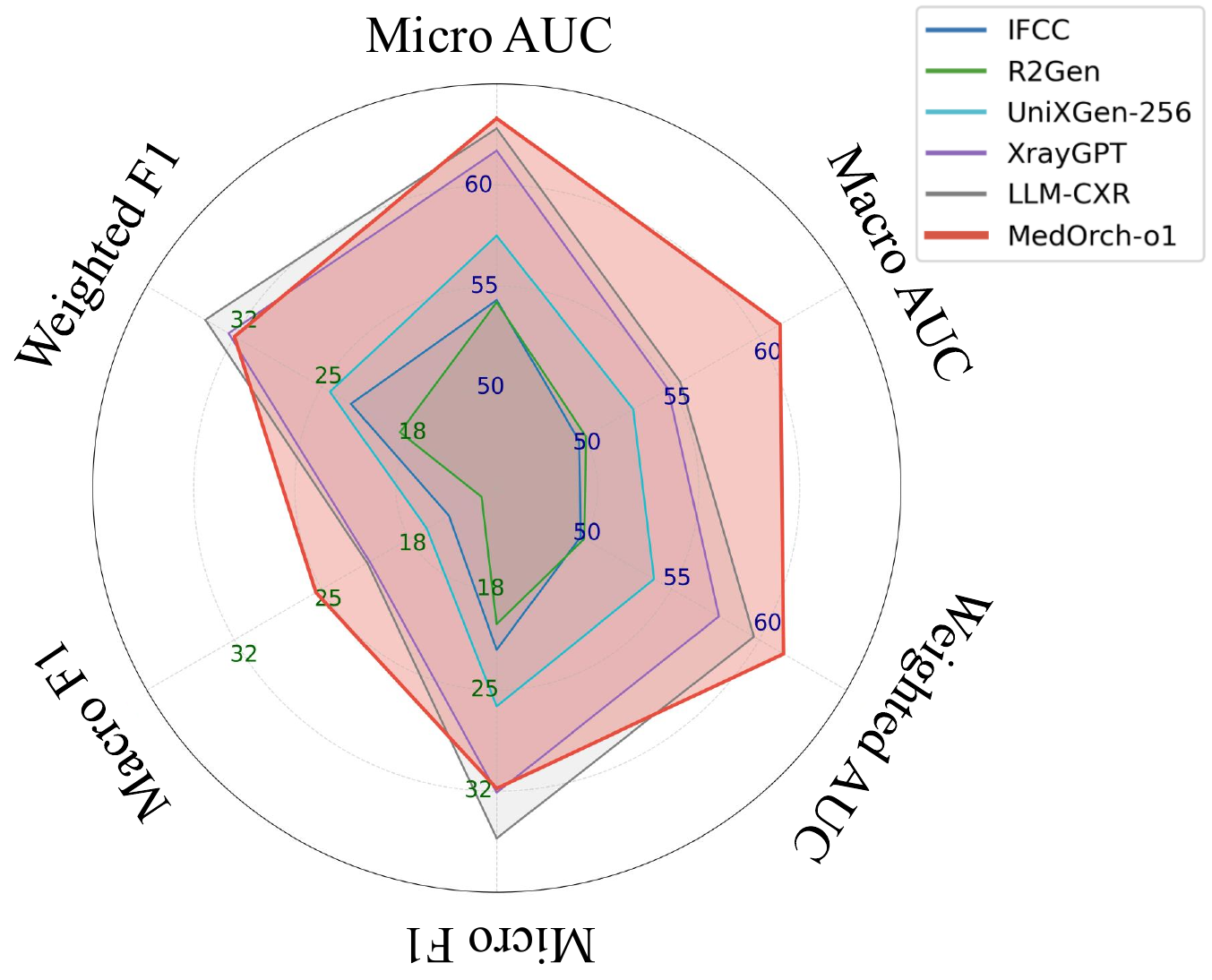}
    \caption{CXR Classification}
    \label{fig:radar-table3}
  \end{subfigure}
  \hfill
  \begin{subfigure}[b]{0.48\textwidth}
    \centering
    \includegraphics[width=\textwidth]{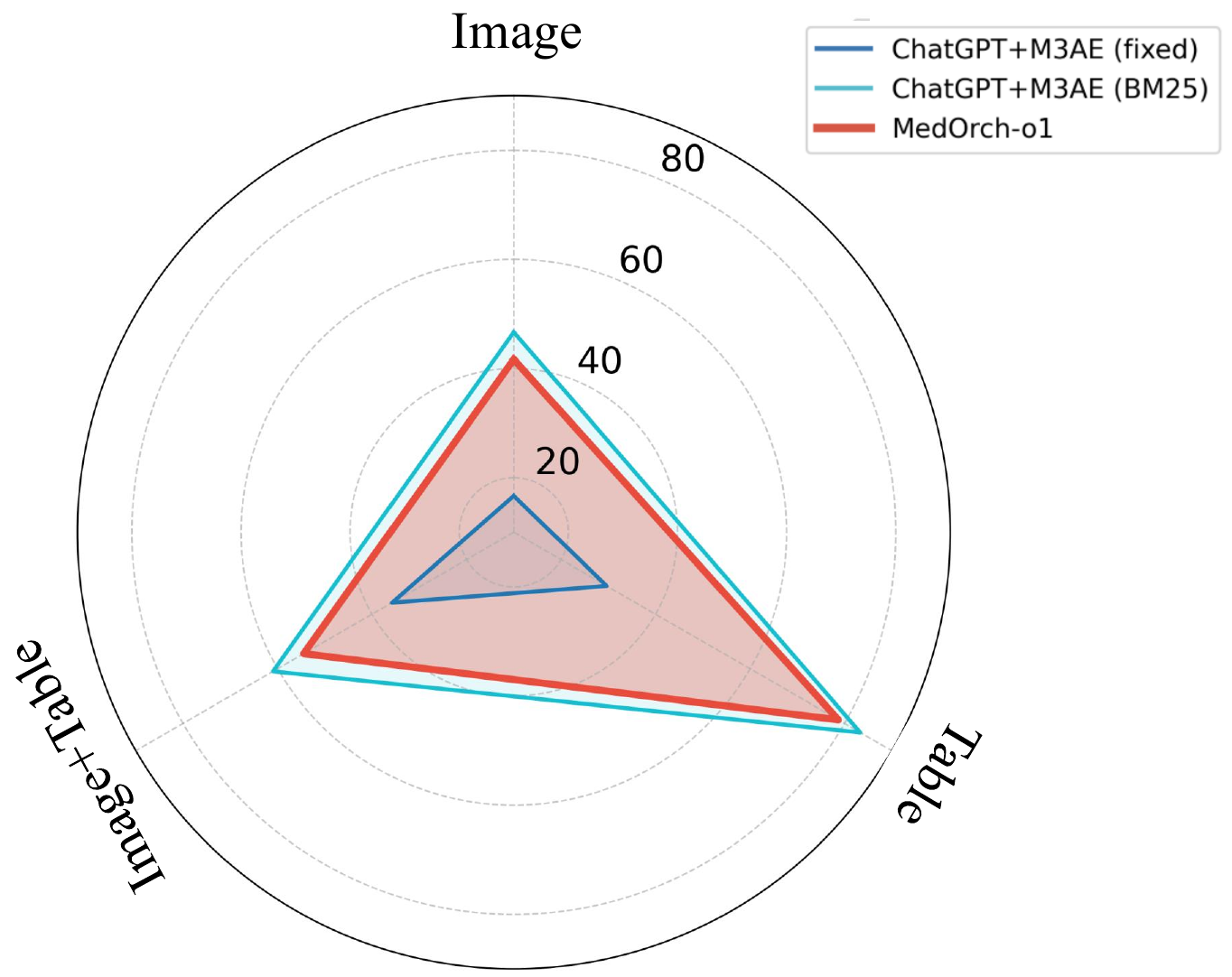}
    \caption{EHR-XQA}
    \label{fig:radar-table4}
  \end{subfigure}

  \caption{Radar charts comparing \name and baselines across medical tasks:
  Alzheimer’s diagnosis (a) and prediction (b),
  chest X-ray classification (c),
  and multimodal question answering (d). \name demonstrates competitive performance across these diverse domains with the same core architecture.}
  \label{fig:radar-2x2}
\end{figure*}

\section{Introduction}

Healthcare decision-making represents one of the most challenging domains for Artificial Intelligence (AI) \cite{singhal2025toward, liu2024survey}. Despite significant advances in medical AI, current systems face three fundamental limitations that hinder their widespread clinical adoption and effectiveness.

\textbf{First, lack of goal-driven, customizable multimodal integration.} Effective clinical decision-making requires adapting the entire reasoning process to a user's specific goal, clinical context, and available data. This involves determining what information is needed, how to decompose complex problems into subtasks, in what order to execute them, and which tools to apply at each stage. Such decisions must account for the heterogeneous nature of clinical information sources, such as patient history, laboratory tests, medical imaging, genetic markers, and clinical literature, each requiring specialized processing \cite{cui2023deep}. However, existing medical AI systems are fundamentally rigid and narrowly focused. While large language models (LLMs) demonstrate remarkable general reasoning capabilities, their effectiveness diminishes when applied in isolation to specialized medical tasks~\cite{kim2404mdagents}. Conversely, specialized models like CheXNet \cite{rajpurkar2017chexnet} excel in specific areas but cannot integrate broader clinical contexts or adapt their reasoning flow. Most of the current medical AI systems require significant re-engineering to adapt to new domains or settings, and lack the ability to incorporate custom agents or tools tailored to local practice patterns or research needs.

\textbf{Second, lack of transparent and auditable reasoning processes.} Many current healthcare AI systems behave like black-box predictors, returning only a final diagnosis without revealing how they reached it (e.g., CheXNet, IDx-DR, and LYNA) \cite{rajpurkar2017chexnet, savoy2020idx, stumpe2018applying}. The absence of intermediate steps makes it difficult for clinicians to trust AI outputs, particularly in high-stakes medical scenarios where understanding the rationale behind a diagnosis is as crucial as the diagnosis itself. Transparent reasoning processes enable healthcare professionals to validate AI reasoning against their clinical expertise, identify potential biases or errors in the analytical pathway, incorporate additional contextual factors that the system may have overlooked. Furthermore, regulatory bodies and medical institutions increasingly demand explainable AI systems that can provide audit trails and justify their decision-making processes \cite{garcia2023functional}.

\textbf{Third, inability to generate diverse diagnostic perspectives.} Current systems typically provide only a single diagnostic conclusion, failing to explore alternative hypotheses or clinical perspectives that could be valuable for complex cases where multiple valid interpretations of clinical data may exist \cite{sadeghi2023brief}. This limitation is particularly problematic in medicine, where considering multiple potential conditions simultaneously is essential for comprehensive patient care. Real-world clinical scenarios often involve ambiguous symptoms, overlapping conditions, or rare diseases that require exploring various diagnostic possibilities \cite{rolando2025labeled}. The inability to generate alternative reasoning strategies, such as reordering steps, varying initial assumptions, or applying different analytical approaches, prevents clinicians from accessing diverse reasoning trajectories that could enhance decision-making in complex cases.

These challenges directly motivate our proposed solution: a framework that orchestrates multiple specialized tools through a reasoning-driven approach that mirrors clinical workflow. We propose \name, a highly customizable system that addresses these limitations. The system dynamically coordinates diverse medical tools across different modalities and domains, and can be easily customized through modular system design to adapt to specific clinical goals, workflows, and institutional needs. It maintains transparent reasoning processes that record the entire decision-making pathway, including tool usage and evidence synthesis. Additionally, it generates diverse reasoning pathways that explore multiple diagnostic trajectories and clinical perspectives for comprehensive decision support.

Our key contributions are summarized as follows:

\begin{enumerate}
\item \textbf{Goal-driven modular multimodal integration.} \name employs a modular, agent-based architecture that enables both autonomous reasoning and flexible customization. The system can autonomously determine what information is needed, decompose complex problems into subtasks, decide execution order, and invoke appropriate tools across diverse modalities, including EMR data, imaging, biomarkers, and literature, during the reasoning process. Simultaneously, the entire workflow can be easily customized through simple configuration changes to align with specific clinical goals, institutional workflows, and practice patterns. The modular design allows new agents or tools to be seamlessly integrated without altering the core reasoning framework, enabling rapid adaptation to different medical specialties and incorporation of custom models tailored to local research needs.

\item \textbf{Transparent reasoning trajectory.} \name system provides complete visibility into what reasoning steps are exposed, how tools are selected and invoked, what evidence each tool returns, and how this evidence is integrated to reach conclusions. This comprehensive audit trail enables healthcare professionals to trace the logical pathway from initial query to final diagnosis, understand the rationale behind each intermediate decision, and identify potential areas for improvement or alternative approaches. Such transparency is crucial for building trust in AI-assisted medical decision-making and meeting regulatory requirements for explainable healthcare AI systems.

\item \textbf{Multiple reasoning trajectories.} \name generates multiple reasoning trajectories per case by varying initial conditions, reasoning strategies, or tool selection approaches, enabling exploration of alternative hypotheses and clinical perspectives. Each trajectory represents a different analytical pathway that may prioritize different types of evidence, apply tools in different sequences, or focus on different diagnostic possibilities. This capability provides clinicians with diverse diagnostic pathways for comprehensive decision support, particularly valuable in complex cases where multiple valid interpretations of clinical data may exist. The system's ability to surface alternative reasoning approaches helps healthcare professionals consider differential diagnoses, validate their clinical intuition, and make more informed decisions by examining the convergence or divergence of different analytical strategies.
\end{enumerate}

\begin{figure}
    \centering
    \includegraphics[width=0.95\textwidth]{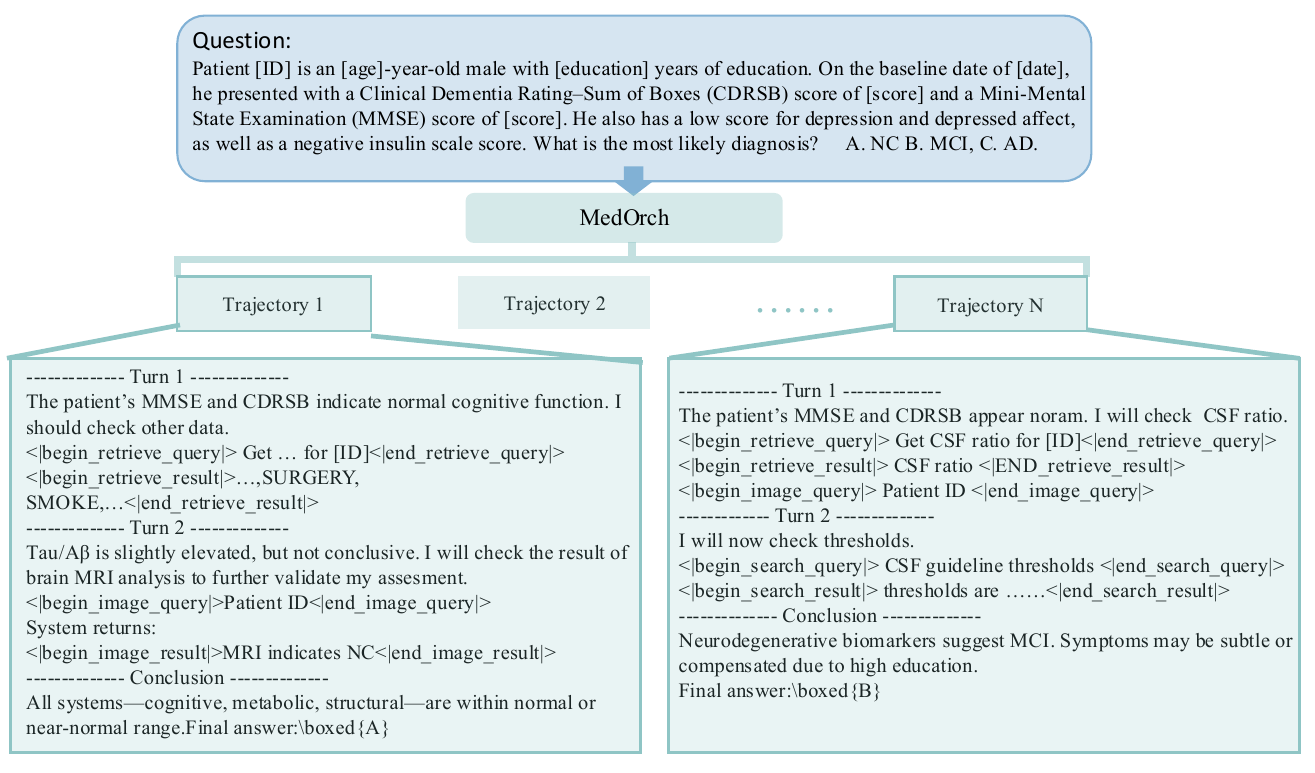}
    \caption{An example of the multiple reasoning trajectories generated by \name, showing how it reasoned and called various tools to solve an Alzheimer's disease diagnosis problem. The entire process is transparent. The intermediate steps, tool invocations, and returned results are all clearly visible. Professionals can review the entire reasoning process.}
    \label{fig:tra}
\end{figure}

To demonstrate \name's core features, we evaluate \name across three distinct medical domains: Alzheimer's disease diagnosis and prediction (ADNI dataset~\cite{petersen2010alzheimer}), chest X-ray interpretation (MIMIC-CXR~\cite{johnson2019mimic}), and multimodal visual question answering (EHRXQA~\cite{bae2023ehrxqa}).
The radar charts in Figure~\ref{fig:radar-2x2} highlight \name's competitive performance across these diverse tasks. On the ADNI dataset, \name achieves 93.26\% accuracy for Alzheimer's diagnosis, outperforming the best baseline by over 4 percentage points, while showing a substantial 19 percentage point improvement in disease progression prediction. In chest X-ray analysis using MIMIC-CXR, \name achieves superior macro-level metrics while maintaining competitive performance on micro-level metrics. For the EHRXQA benchmark, \name achieves strong performance across table-based, image-based, and image+table questions without any manually crafted task-specific training data.
These evaluation results demonstrate \name's competitive performance and flexibility, achieved by seamlessly integrating domain-appropriate tools without architectural changes. 
Besides, as Figure \ref{fig:tra} shows, \name provides transparent reasoning processes that record all intermediate steps, tool invocations, and returned results, to enhance the trust and understanding of healthcare professionals. The performance improvement from multiple reasoning trajectories, as shown in Table \ref{table1} and \ref{table2}, demonstrates the significant value of providing multiple reasoning trajectories to support clinical decision-making.
These findings suggest that \name provides a generalizable and practical foundation for next-generation medical AI systems that are not only accurate and transparent but also deeply adaptable to real-world clinical needs.

\section{Related Work}

\subsection{AI Systems for Medical Decision Support}

\textbf{Medical Language Models.} Recent years have seen significant advances in language models specialized for healthcare applications. Med-PaLM \cite{singhal2025toward} demonstrated expert-level performance on medical licensing exams but operates primarily in the text domain without integrating multimodal clinical data. Similarly, GatorTron \cite{yang2022large} leveraged massive clinical text corpora to enhance medical language understanding, while Clinical-BERT \cite{alsentzer2019publicly} adapted BERT specifically for clinical text. BioMedLM \cite{bolton2023biomedlm} focused on biomedical literature understanding. These models show impressive medical knowledge but lack mechanisms for autonomous tool use or integration of non-textual data.

\textbf{Clinical Decision Support Systems.} Traditional clinical decision support systems like DXplain \cite{barnett1987dxplain} and Isabel \cite{ramnarayan2006isabel} employ rule-based or statistical approaches to diagnostic assistance. More recent systems like Watson for Oncology \cite{yu2021early} incorporate machine learning but typically operate in narrowly defined domains with predefined information pipelines rather than dynamically integrating diverse data types. 

\textbf{Medical Vision-Language Models.} Several studies have explored integrated models for medical imaging and text. CLIP-based models adapted for medical imaging \cite{wang2022medclip} demonstrated strong representation learning but lack reasoning capabilities across diverse clinical contexts. RadFM \cite{wu2023towards} specialized in radiology report generation but cannot generalize to other medical domains.
MultiMedBench \cite{tu2024towards} introduced benchmarks for multimodal medical AI across various tasks but focused on standalone performance rather than integrated reasoning. MedVInT \cite{zhang2024pmcvqavisualinstructiontuning} achieved strong performance on medical visual question answering tasks but lacks the ability to extend to other non-visual clinical data crucial for comprehensive diagnosis.

\subsection{Tool-Using and Agentic AI Systems}

\textbf{LLMs with Tool Use.} Recent work has demonstrated the value of equipping language models with external tools. Early approaches such as Toolformer \cite{schick2023toolformerlanguagemodelsteach} and HuggingGPT \cite{shen2023hugginggptsolvingaitasks} laid the groundwork: Toolformer showed that models can learn to use APIs through self-labeled in-context examples, while HuggingGPT proposed a framework for coordinating multiple specialized models. However, these systems typically rely on predefined or static tool invocation strategies rather than reasoning-driven decision making. Subsequent efforts like Gorilla \cite{patil2024gorilla}, ToolACE \cite{liu2024toolace}, and xLAM \cite{zhang2024xlam,prabhakar2025apigen} demonstrated that fine-tuned LLMs can generalize to invoke previously unseen tools and solve complex tasks across multi-turn dialogues. While effective, these works primarily focus on the correctness of tool calls rather than the model’s ability to reason with tools in a dynamic and compositional manner. More recently, the emergence of large reasoning models (LRMs) has led to a new wave of systems, such as Search-o1 \cite{li2025search}, WebThinker \cite{li2025webthinker}, and Agentic Reasoning \cite{wu2025agentic}, that show strong zero-shot performance on complex, tool-augmented reasoning tasks. Although these systems provide the foundation for our work, they often underexplore the role of specialized tools in addressing the most challenging real-world problems such as medical diagnosis.

\textbf{Multi-Agent and Reasoning Systems.} The concept of multi-agent systems has gained traction in AI research. AutoGen \cite{wu2023autogenenablingnextgenllm} proposed a framework for multi-agent conversation, while Reflexion \cite{shinn2023reflexionlanguageagentsverbal} incorporated reflection mechanisms to improve reasoning. These frameworks demonstrate the potential of tool-using agents but lack the domain-specific adaptations necessary for medical applications, where specialized tools and multimodal integration are crucial for effective reasoning across complex clinical scenarios.

\section{Proposed Method}

\subsection{System Overview}
\begin{figure}
    \centering
    \includegraphics[width=0.95\textwidth]{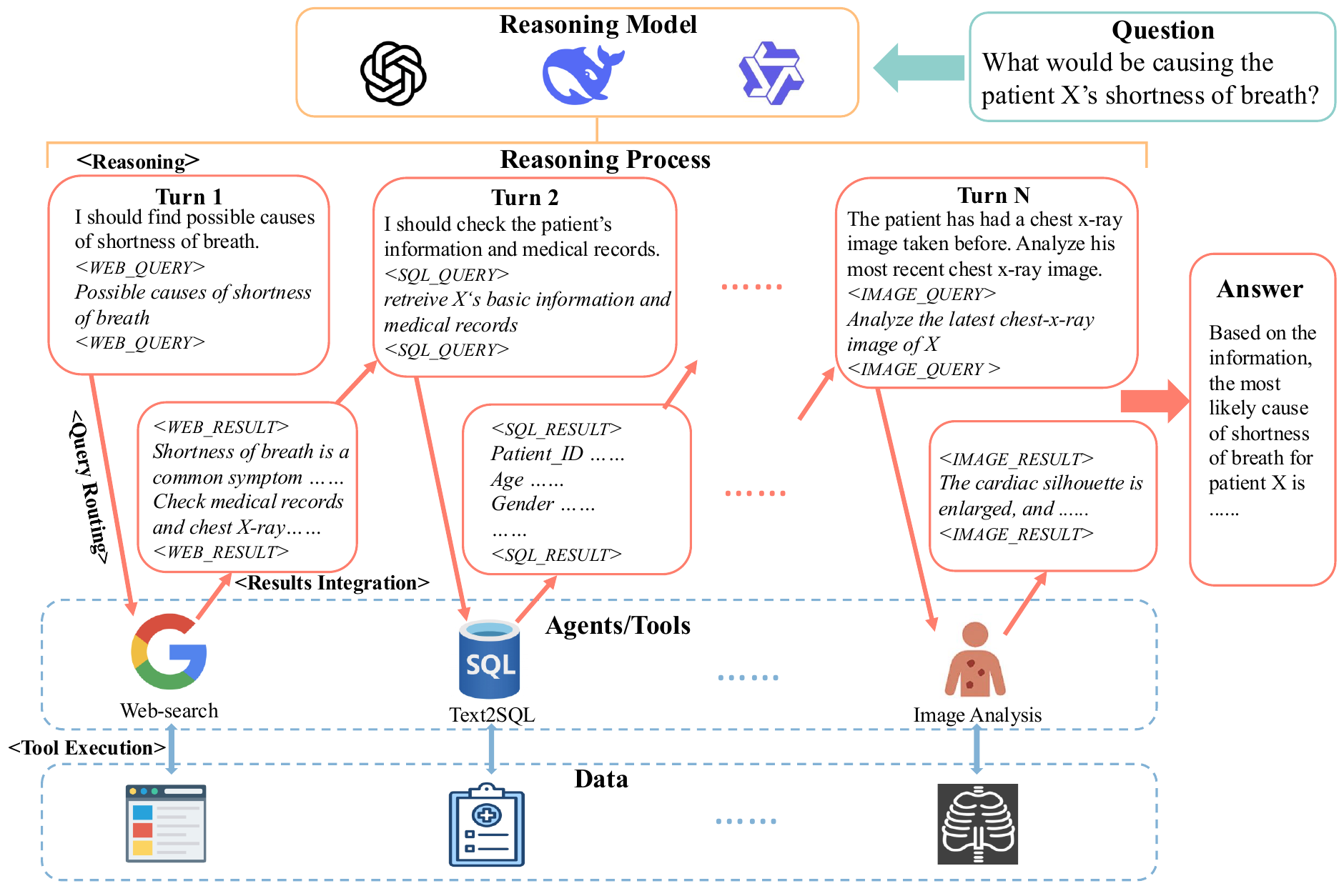}
    \caption{The overall architecture of \name. The LLM maintains a chain-of-thought and emits tool-call tokens whenever external evidence is required. In the particular example shown in the figure, the reasoning model (1) first queries a web-search agent for differential causes of dyspnea; (2) then retrieves the patient’s demographics and history through a Text2SQL agent; (3) identifies gaps in the available evidence and proceeds with further tool-based reasoning; (4) consults a chest X-ray image-analysis agent, whose findings serve as the final piece in resolving the diagnostic case; and (5) concludes by summarizing the key evidence and delivering a final diagnosis. Each agent’s output is seamlessly integrated back into the model’s reasoning trace, enabling transparent, multi-step deliberation until a final answer is produced.}
    \label{fig:architecture}
\end{figure}

\name is a goal-driven framework that dynamically orchestrates specialized tools and agents for comprehensive medical decision support. Unlike conventional medical AI systems that are rigid and narrowly focused, \name provides a modular, customizable architecture that enables both autonomous reasoning and flexible adaptation. The system integrates tool utilization directly into the reasoning process, allowing the model to determine what information is needed, how to decompose problems into subtasks, the execution order, and which specialized capabilities to invoke as the analysis unfolds. Crucially, the entire workflow can be easily customized through simple configuration changes to align with different clinical workflows, institutional practices, and specialized requirements.

Figure~\ref{fig:architecture} illustrates \name's overall architecture. At the core is a reasoning model that can invoke specialized agents during clinical analysis. The system maintains a tool registry that contains structured representations of all available tools, their capabilities, and appropriate usage contexts. This registry is injected into the reasoning model's context during initialization, enabling it to reason about and utilize tools without requiring retraining. Each tool entry includes semantic descriptions, input/output specifications, and example usage patterns. New tools can be added through the registry system without architectural modifications, demonstrating the framework's modular extensibility for emerging medical technologies and institutional requirements.

\subsection{Core Reasoning Mechanism}
The reasoning process in \name is formulated as a sequence of inference steps that dynamically incorporates tool invocation and external information integration. Formally, this process can be defined as:
\begin{equation}
P(r, a | q, e, k) = \prod_t P(r_t | r_{<t}, q, e_{\leq t}, k_{\leq t}) \times \prod_t P(a_t | a_{<t}, r, q, e, k),
\end{equation}
where $q$ represents the medical query or task, $r$ denotes the reasoning chain, $a$ is the final answer, $e$ represents tool-generated outputs, and $k$ contains knowledge. The subscript $t$ indicates the token position in the reasoning sequence.

Central to this reasoning framework is the generation of specialized tool-calling tokens within the model's reasoning process. These tokens, such as \texttt{[IMAGE\_QUERY]}, \texttt{[SQL\_QUERY]}, or \texttt{[WEB\_QUERY]}, signal moments when external information or specialized analysis would strengthen the reasoning. Importantly, these tokens are generated by the reasoning model itself during inference, without any fine-tuning or additional training. Instead, the model learns to produce these tokens through in-context learning via detailed instructions and examples in its system prompt, which comprehensively describes how to invoke various tools using these special tokens.

This in-context learning approach enables the model to dynamically determine when specialized tools are needed based solely on the requirements of the current reasoning task and the information available. For instance, when analyzing patient data for Alzheimer's disease risk, the model might autonomously generate a \texttt{[IMAGE\_QUERY]} token to trigger brain MRI analysis at the precise point in its reasoning where visual information would resolve uncertainty or provide critical diagnostic evidence.

Upon the model generating such a token, the reasoning process follows a three-step protocol:
\begin{enumerate}
    \item \textbf{Query Routing} The system extracts from the reasoning chain the specific query and relevant context, which are then routed to an appropriate agent or tool.
    \item \textbf{Tool Execution:} The selected tool processes the query and generates outputs.
    \item \textbf{Results Integration:} These outputs are seamlessly reintegrated into the ongoing reasoning flow, becoming available as context for subsequent reasoning steps.
\end{enumerate}

The reasoning process continues until the model determines it has sufficient information to provide a comprehensive answer, indicated by the generation of a conclusion without further tool-calling tokens. 

While the above describes the default autonomous reasoning behavior, \name's modular architecture enables easy customization of the reasoning workflow through configuration modifications. Users can specify preferred reasoning strategies, such as prioritizing certain types of evidence, enforcing specific tool invocation sequences, or requiring particular information gathering steps for specific clinical scenarios. These customizations are implemented through modified system prompts and tool registry configurations, allowing the framework to adapt to institutional protocols, clinical guidelines, or specialized diagnostic workflows without requiring model retraining or architectural changes.

We give a simple example to show how \name works in Figure~\ref{fig:architecture}. When assessing potential Alzheimer's disease, the reasoning flow unfolds organically: the model first queries web-search agent for diagnostic guidelines to establish clinical frameworks, then activates Text2SQL to retrieve patient's information. 
Then it identifies gaps in the available evidence and proceeds with further tool-based reasoning. Finally, It consults a chest X-ray image-analysis agent, whose findings serve as the final piece in resolving the diagnostic case. So, it concludes by summarizing the key evidence and delivering a final diagnosis. Each agent’s output is seamlessly integrated back into the model’s reasoning trace, enabling transparent, multi-step deliberation until a final answer is produced.

The system can generate diverse reasoning pathways for the same case by varying analytical priorities or diagnostic hypotheses, providing clinicians with complementary perspectives for complex cases where multiple valid interpretations may exist.

\subsection{General-Purpose Agents}

\name is built upon a set of general-purpose agents that form the foundation of its problem-solving capabilities. These agents enable broad reasoning functionality across diverse tasks. 

\textbf{Web-search Agent.} Searches the internet to retrieve relevant medical information across a broad range of sources. This agent extends the reasoning model's knowledge by accessing recent medical literature, clinical discussions, case reports, health statistics, and emerging treatment approaches that may not be present in the model's training corpus. For instance, when investigating a rare condition or evaluating current treatment protocols, it can find the most up-to-date information published online, allowing the reasoning model to incorporate recent developments into its analysis.

\textbf{Coding Agent.} Executes programmatic tasks through dynamic code generation and execution based on specific reasoning requirements. This agent enables the system to perform complex data processing, statistical analysis, algorithmic implementations, and custom computational workflows as needed during clinical reasoning. It can develop and run code to analyze medical data, implement disease progression models, generate visualizations of patient data, process structured medical information, and execute specialized clinical algorithms.

\textbf{Text2SQL Agent.} Transforms natural language queries about patient information into structured database queries, enabling the retrieval of relevant clinical data from electronic health records. This agent allows the reasoning model to access specific patient parameters, test results, and medical history. Additionally, it can retrieve similar patient cases in the clinical database based on demographic features, symptom patterns, or diagnostic criteria, providing valuable reference cases to inform the current diagnostic process. These comparative cases often serve as empirical evidence to support diagnostic hypotheses or treatment considerations.

\textbf{Retrieval Agent.} A retrieval-augmented generation (RAG) system containing structured representations of clinical practice guidelines from authoritative medical organizations. This knowledge base provides the reasoning model with access to standardized diagnostic criteria, treatment protocols, and best practices that represent the medical consensus for various conditions.

\subsection{Task-Specific Medical Agents}
 On top of the aformentioned general-purpose agents, \name integrates domain-specific medical agents tailored to specialized modalities and knowledge areas, allowing the system to achieve strong performance on complex clinical challenges. Specifically, we include the following specialized agents designed for specific medical domains and data modalities.

\textbf{Medical Image Analysis Agent.} Processes medical images through a combination of neural networks and specialized analysis tools. This agent employs pre-trained neural networks for image feature extraction and classification tasks across various imaging modalities. For example, MRI analysis in Alzheimer's disease diagnosis interfaces with FreeSurfer, an open-source suite widely used for automated brain segmentation and surface-based morphometry~\cite{fischl2012freesurfer}, to perform automated segmentation and volumetric analysis of brain structures, extracting measurements such as hippocampal volume, cortical thickness, and ventricular size. It also employs neural networks to extract features and perform the classification task to provide references for the reasoning process. The agent quantifies these neuroanatomical features and converts them into structured numerical data that the reasoning model can directly incorporate into its diagnostic assessment. This provides objective biomarkers that support clinical reasoning.

\textbf{Medical Imaging QA Agent.} A multimodal model fine-tuned on medical imaging question-answering datasets that can interpret various medical images and respond to queries about them. This agent processes radiological images, pathology slides, dermatological photographs, and other visual medical data to extract clinically relevant information based on natural language questions. This agent provides detailed narrative descriptions, answers specific queries about anatomical structures, discusses differential diagnoses based on visual findings, and explains the clinical significance of observed features. It handles complex questions requiring integration of visual features with medical knowledge, such as "What is the likely etiology of the opacity in the right lower lobe of this chest X-ray?". This capability is particularly valuable for bringing radiological insights directly into the clinical reasoning process.

\textbf{Clinical Knowledge Graph Agent.} Constructs and maintains a structured knowledge graph representing the relationships between clinical entities (symptoms, conditions, treatments) mentioned during reasoning. This agent enables the system to track logical connections between medical concepts and maintain coherence during extended reasoning chains. The knowledge graph functions as an external memory that can be queried to retrieve previously established relationships or to identify potential diagnostic patterns, particularly valuable for complex cases involving multiple interrelated symptoms or conditions.

\textbf{Longitudinal Data Analysis Agent.} 
Specializes in processing and interpreting time-series medical data to identify clinically significant patterns and trends
This agent ingests irregular, multivariate time-series records, like laboratory values, vital signs, cognitive scores, or imaging biomarkers collected at successive visits, and converts them into information the reasoning model can use. After aligning observations to a common timeline and filling small gaps with simple interpolation, it derives trend descriptors such as moving averages, slopes, and rate-of-change statistics that summarize clinically relevant progression. By supplying this information to the reasoning process, \name can reason over a patient’s longitudinal history rather than isolated snapshots.

The modular architecture of \name enables seamless integration of additional specialized agents as clinical needs evolve. New tools can be easily incorporated without modifying the core reasoning engine, such as drug interaction checking agents leveraging databases like DrugBank~\cite{knox2024drugbank} or decision support tools like the Framingham Risk Score calculators~\cite{hemann2007framingham}. The tool registry system automatically makes these new capabilities available to the reasoning model through simple configuration updates, allowing \name to adapt to diverse clinical environments and emerging medical technologies without requiring system-wide retraining.

\section{Experiments}
\subsection{Experiment Overview}
\begin{table}[t]
\centering
\caption{Comparison of methods on Alzheimer’s disease diagnosis tasks. \name (o1-mini) with a best@5 strategy achieves the highest accuracy of 93.26\% on the three-class classification task, while \name (GPT-4o) with best@5 attains the best performance of 89.50\% on the four-class task. Accuracy improves from 88.71\% (best@1) to 93.26\% (best@5) in the three-class setting, with similar gains observed in the four-class case, demonstrating the advantage of generating multiple reasoning trajectories.}
\setlength{\tabcolsep}{1.8pt}
\renewcommand{\arraystretch}{1}
\begin{tabular*}{\textwidth}{@{\extracolsep{\fill}} l|
  >{\centering\arraybackslash}p{1.2cm}
  >{\centering\arraybackslash}p{1.5cm}
  >{\centering\arraybackslash}p{1.5cm}|
  >{\centering\arraybackslash}p{1.5cm}
  >{\centering\arraybackslash}p{1.5cm}
  >{\centering\arraybackslash}p{1.5cm}
}
\toprule
\textbf{Method} & \multicolumn{3}{c|}{\textbf{AD vs. MCI vs. NC}} & \multicolumn{3}{c}{\textbf{AD vs. EMCI vs. LMCI vs. NC}}  \\
 & ACC & SPE & SEN & ACC & SPE & SEN \\
\midrule
CNN~\cite{huang2018denselyconnectedconvolutionalnetworks} & 61.12 & 60.85 & 59.34 & 58.37 & 54.10 & 51.08 \\
U-Net~\cite{fan2021u} & 87.65 & - & - & 86.47 & - & - \\
slice attention module~\cite{{huo2022multistage}} & 78.90 & 73.33 & 91.10 & 87.50 & \textbf{95.60} & 63.33 \\
Hypergraph Diffusion~\cite{aviles2022multi} & 83.75 & 80.64 & 83.07 & 86.47 & 78.52 & 82.16 \\
M3AE~\cite{chen2022m3ae} & 60.18 & 54.28 & 58.24 & 53.45 & 54.01 & 42.37 \\
ConvNext+GPT~\cite{feng2023largelanguagemodelsimprove} & 89.05 & 87.33 & 89.29 & 87.63 & 85.79 & 87.25 \\
\midrule
\textbf{\name(GPT-4o, best@1)} & 84.95 & 85.54 & 84.60 & 83.21 & 80.14 & 83.45 \\
\textbf{\name(o1-mini, best@1)} & 88.71 & 85.67 & 90.32 & 86.38 & 84.33 & 84.25 \\
\midrule
\textbf{\name(GPT-4o, majority@5)} & 86.68 & 84.45 & 86.92 & 84.56 & 83.38 & 83.25 \\
\textbf{\name(o1-mini, majority@5)} & 89.93 & 89.03 & 91.45 & 86.84 & 86.28 & 88.21 \\
\midrule
\textbf{\name(GPT-4o, best@5)} & 91.22 & \textbf{90.84} & 91.28 & \textbf{89.50} & 92.36 & \textbf{88.84} \\
\textbf{\name(o1-mini, best@5)} & \textbf{93.26} & 88.37 & \textbf{92.35} & 88.24 & 89.34 & 88.65 \\
\bottomrule
\end{tabular*}
\label{table1}
\end{table}

To demonstrate the practical impact of the \name framework, we evaluate it on three distinct medical tasks, highlighting different core capabilities of the system.

First, in the ADNI setting~\cite{petersen2010alzheimer}, \name must synthesize heterogeneous inputs including demographic variables, cognitive assessments, fluid biomarkers, and brain MRIs, to perform both diagnostic classification and disease progression prediction. This task highlights \name’s ability to modularly integrate multimodal tools and explore diverse reasoning trajectories. By prompting \name to solve the same case multiple times, we obtain a set of reasoning traces that reflect different diagnostic perspectives. This diversity leads to statistically significant gains in accuracy and is particularly useful in decision-support contexts, where surfacing alternative, evidence-based hypotheses benefits clinical judgment.

Second, on the MIMIC-CXR dataset, \name identifies chest X-ray abnormalities and incorporates structured patient history when available. This experiment illustrates \name’s extensibility: new image encoders can be plugged into the system without retraining the core reasoning engine. Without any modification to the core orchestration architecture, we observe state-of-the-art macro metrics while maintaining competitive micro metrics.

Finally, the EHRXQA benchmark poses clinical questions that require reasoning over structured EHR data and visual inputs. Here, \name dynamically switches between symbolic querying, visual interpretation, and natural language explanation within a single dialogue. This task showcases \name’s ability to coordinate multiple domain-specific agents in a seamless, mixed-modality workflow, without requiring any changes to the underlying reasoning architecture.

Across all three tasks, we retain a fixed reasoning model and simply register the appropriate tools, such as a Text2SQL agent for querying structured records or a medical-VQA agent for image understanding, via the existing interface. Compared to baselines optimized for individual domains, \name achieves competitive performance while offering a transparent, auditable, and flexible reasoning process that supports clinical trust and adaptability.

\subsection{Alzheimer's Disease Assessment}

\begin{table}[t]
\centering
\caption{Comparison of models on prediction performance. \name (o1-mini) with best@5 reaches the best accuracy (50.4\%). \name (o1-mini) with achieves a best@1 accuracy of 38.61\%, which is about 7 percentage points higher than the best baseline. The results demonstrate the superiority of \name on tasks that require complex reasoning and integration of multiple multimodal data. Notably, when selecting the best result from $5$ independently generated reasoning trajectories (best@5), accuracy increases to 50.35\%, highlighting the value of exploring multiple diagnostic paths. } 
\begin{tabular}{l|ccc}
\toprule
\textbf{Method} & \multicolumn{3}{c}{\textbf{AD vs. MCI vs. NC}} \\
& \textbf{Acc} & \textbf{SPE} & \textbf{SEN} \\
\midrule
M3AE~\cite{chen2022m3ae} & 25.31 & 25.05 & 25.80 \\   
ConvNext+GPT~\cite{feng2023largelanguagemodelsimprove} & 31.01 & 28.76 & 32.18 \\
\midrule
\textbf{\name (GPT-4o, best@1)} & 37.97 & \textbf{35.55} & 38.51 \\
\textbf{\name (o1-mini, best@1)} & \textbf{38.61} & 34.65 & \textbf{38.80} \\
\midrule
\textbf{\name (GPT-4o, majority@5)} & 40.51 & 41.53 & 39.24 \\
\textbf{\name (o1-mini, majority@5)} & 39.87 & 38.67 & 38.45 \\
\midrule
\textbf{\name (GPT-4o, best@5)} & 50.28 & 47.25 & \textbf{52.35} \\
\textbf{\name (o1-mini, best@5)} & \textbf{50.35} & \textbf{48.82} & 52.21 \\
\bottomrule
\end{tabular}
\label{table2}
\end{table}
\subsubsection{Dataset and Task}

The ADNI~\cite{petersen2010alzheimer} dataset is a longitudinal study designed to track the progression of Alzheimer's disease. We utilize data from ADNI-1, ADNI-2, which originally categorizes patients into five diagnostic groups: Cognitively Normal (CN), Significant Memory Concern (SMC), Early Mild Cognitive Impairment (EMCI), Late Mild Cognitive Impairment (LMCI), and Alzheimer's Disease (AD). For our primary analyses, we group these into three main diagnostic categories: Normal Control (NC, including CN and SMC), Mild Cognitive Impairment (MCI, including EMCI and LMCI), and Alzheimer's Disease (AD). 
The dataset provides rich multimodal information, including demographic information (age, gender, education level, etc.), neuropsychological assessments (MMSE, ADAS-Cog, CDR, etc.), MRI scans, PET imaging data, genetic information (APOE genotype), cerebrospinal fluid (CSF) biomarkers, and longitudinal patient data spanning multiple visits.
This comprehensive dataset provides an ideal testbed for evaluating \name's ability to integrate multimodal information and reason across diverse clinical data types in a neurological disease context.

Based on the ADNI dataset, we formulate two clinically meaningful question-answering tasks:

\textbf{Diagnostic QA Task.} For this task, we create questions that provide partial patient information (an identifier along with a limited random selection of available data, which might include some demographics, biomarkers, or cognitive scores) and require the system to determine the patient's diagnostic status (NC, MCI, or AD). This partial-information approach simulates real clinical scenarios where initial data is incomplete. To ensure successful diagnosis, the system must autonomously determine what additional information is required, invoke appropriate tools to retrieve relevant patient data, comprehensively analyze cognitive assessments, neuroimaging features, and biomarkers, and ultimately synthesize these data to reach a diagnostic conclusion. For more fine-grained analyses, we also conduct experiments for a more granular four-class setting.

A typical diagnostic question format is:

\begin{tcolorbox}[
  colframe=blue!60,   
  colback=blue!5,     
  boxrule=0.8pt,      
  arc=2mm,            
  left=2mm, right=2mm, top=1mm, bottom=1mm, 
  enhanced             
]
\textit{Patient [ID] is a [age]-year-old man with [education] years of education.  
On the date of [date], he presented with a Clinical Dementia Rating--Sum of Boxes (CDRSB) score of [score]  
and a Mini-Mental State Examination (MMSE) score of [score].  
He also has [additional information].  
Given these clinical parameters, what is the most likely diagnosis for this patient at this point in time?}
\end{tcolorbox}

We generate a comprehensive evaluation dataset consisting of 638 diagnostic QA examples to thoroughly assess the system's diagnostic capabilities across a wide range of patient presentations and clinical parameters.

\textbf{Predictive QA Task.} For this task, we present historical patient data from earlier visits and ask the system to predict the current diagnostic status. This requires analyzing disease progression trajectories and identifying patterns indicative of stability or deterioration. A typical predictive question format is:

\begin{tcolorbox}[
  colframe=blue!60,   
  colback=blue!5,     
  boxrule=0.8pt,      
  arc=2mm,            
  left=2mm, right=2mm, top=1mm, bottom=1mm, 
  enhanced             
]
\textit{Patient [ID] had [previous diagnostic status] at his visit [timepoint]. Here is some clinical information and corresponding visit time: [timepoint]:[some clinical information]. You can also query the database for more information or visits.   Based on his data, predict his latest [timepoint] diagnostic classification.}
\end{tcolorbox}

For the predictive task, we generate 158 questions that challenge the system to analyze temporal patterns and disease progression trajectories, representing a diverse set of progression scenarios from stability to rapid decline.

\subsubsection{Baselines}

To evaluate \name's performance on these tasks, we compare it against the following baselines.

\textbf{Baseline CNN} \cite{huang2018denselyconnectedconvolutionalnetworks}: A basic CNN approach using DenseNet architecture for feature extraction from neuroimaging data.
    
\textbf{U-Net} \cite{fan2021u}: A U-shaped network architecture adapted for AD classification.
    
\textbf{Slice Attention Module} \cite{huo2022multistage}: A 3D abnormal perception depth residual network that integrates squeeze‑and‑excitation residual blocks (RSE) with a recurrent slice attention (RSA) module to capture both channel importance and long‑range dependencies across MRI slices for multistage Alzheimer’s diagnosis.

\textbf{Hypergraph Diffusion} \cite{aviles2022multi}: A semi-supervised multi-modal hypergraph framework that combines a dual perturbation-invariant embedding strategy with a dynamically adjusted diffusion model to enhance early Alzheimer’s diagnosis using both imaging and non-imaging data.

\textbf{M3AE} \cite{chen2022m3ae}: A self‑supervised model that jointly masks and reconstructs image patches and text tokens to learn transferable cross‑modal representations, particularly effective for medical vision‑language tasks.

\textbf{ConvNext+GPT} \cite{feng2023largelanguagemodelsimprove}: A recent state-of-the-art approach that integrates the ConvNeXt vision encoder with GPT for processing non-imaging data, representing one of the first applications of large language models to AD diagnosis.

Following the evaluation approach established in previous work, we evaluate performance using three standard metrics: \textbf{ACC} (Accuracy), \textbf{SPE} (Specificity), \textbf{SEN} (Sensitivity).

\subsubsection{Results and Analysis}

We implement \name with two different reasoning models: GPT-4o and o1-mini. They serve as the core reasoning engines that orchestrate tool calling and integrate information during the diagnostic process. To evaluate the framework's performance under different conditions, we employ three evaluation strategies:
\textbf{best@1}: The model makes a single attempt to answer the question. 
\textbf{majority@5}: The model generates five different reasoning trajectories for the same question, and the final diagnosis is determined by majority voting among these paths. 
\textbf{best@5}: The model generates five different reasoning paths, and if any of them produce the correct diagnosis, it is counted as correct.

Table \ref{table1} presents the comparative performance of \name and baseline methods on the diagnostic classification tasks, while Table \ref{table2} shows results for the more challenging prediction task. Several key findings emerge from these results.

\textbf{Diagnostic Performance.} On the three-class AD vs. MCI vs. NC classification task, \name with o1-mini using the best@5 strategy achieves the highest accuracy of 93.26\%, outperforming all baseline methods, including the previous state-of-the-art ConvNext+GPT approach (89.05\%). This represents a 4.21\% absolute improvement in accuracy. Similar patterns are observed in the four-class classification tasks (AD vs. EMCI vs. LMCI vs. NC), where \name consistently outperforms baselines across all metrics. The significant improvement highlights the value of \name' dynamic tool-calling approach, which can adapt its information-gathering strategy to each specific case.

\textbf{Prediction Performance.} The disease progression prediction task proves more challenging for all. However, \name shows a more pronounced advantage in this setting, achieving 50.35\% accuracy with the best@5 and 38.61\% with best@1. This significant lead over the baselines highlights \name' superior ability to analyze temporal patterns in patient data and identify relevant progression indicators through its specialized tools and agents.

\textbf{Effect of Multiple Reasoning Trajectories.}
Table \ref{table1} reports \name’s accuracy under different evaluation strategies. Accuracy rises from 88.71\% (\textit{best@1}) to 93.26\% (\textit{best@5}) on the three-class diagnosis task, with analogous gains in the four-class setting. The results in Table \ref{table2} also show a similar accuracy improvement. The accuracy gains confirm that prompting \name to answer the same clinical question multiple times, thus producing a set of distinct reasoning traces, significantly improves diagnostic performance. This multi-answer strategy is especially valuable for professionals, as it offers several evidence-based perspectives while maintaining complete transparency: each step of the logic is logged, together with the tools or agents invoked and the exact outputs they return. An example of multiple transparent reasoning trajectories is provided in Appendix \ref{sec:mult_tra}.

\subsection{Chest X-ray Diagnosis}

\subsubsection{Dataset and Task}

The MIMIC-CXR dataset~\cite{johnson2019mimic} is a large, publicly available collection of chest radiographs with associated radiology reports. The dataset contains 377,110 images from 227,835 radiographic studies performed on 65,379 patients. Each study includes chest X-ray images (frontal and lateral views when available) along with corresponding free-text radiology reports written by board-certified radiologists.

We formulate a chest X-ray diagnostic task to evaluate \name' ability to analyze radiological images and provide accurate clinical assessments. For this task, the system is required to examine chest X-ray images and determine the presence of the 14 thoracic pathologies labeled in the MIMIC-CXR dataset.

\subsubsection{Baselines}

We compare \name with several competitive multimodal baselines:

\textbf{IFCC}~\cite{delbrouck2022improving}: A method that incorporates intermediate feature conditioning for radiology generation and understanding.

\textbf{R2Gen}~\cite{chen2020generating}: A transformer-based report generation model trained on paired image-text data.

\textbf{UniXGen}~\cite{lee2023vision}: A bidirectional transformer model trained from scratch on image-text pairs.

\textbf{XrayGPT}~\cite{thawkar2023xraygpt}: A Vicuna-based LLM equipped with a MedCLIP vision encoder.
    
\textbf{LLM-CXR}~\cite{lee2024llmcxrinstructionfinetunedllmcxr}: An instruction-tuned LLM trained with vision-language alignment via clinical tokenization.

\begin{table}[t]
\centering
\caption{Comparison of models on MIMIC-CXR classification and detection performance. Metrics follow prior work. \name (o1-mini) achieves the highest scores in every AUC column and in macro-F1. \name (GPT-4o) also achieves competitive performance compared to the best baseline. The results demonstrate that \name’s can be easily adapted to different domains by integrating new agents or tools.}
\begin{tabular}{l|ccc|ccc}
\toprule
\textbf{Method} & \multicolumn{3}{c|}{\textbf{AUC}} & \multicolumn{3}{c}{\textbf{F1}}\\
& \textbf{Micro} & \textbf{Macro} & \textbf{Weighted}  & \textbf{Micro} & \textbf{Macro} & \textbf{Weighted}\\
\midrule
IFCC~\cite{delbrouck2022improving} & 54.3 & 49.7 & 49.8 & 22.0 & 14.1 & 22.5 \\
R2Gen~\cite{chen2020generating} & 54.2 & 50.1 & 50.0 & 20.1 & 11.3 & 18.3 \\
UniXGen-256~\cite{lee2023vision} & 57.5 & 52.8 & 54.0 & 26.2 & 16.0 & 24.3 \\
XrayGPT~\cite{thawkar2023xraygpt} & 61.7 & 54.8 & 57.7 & 32.6 & 20.9 & 33.0 \\
LLM-CXR~\cite{lee2024llmcxrinstructionfinetunedllmcxr} & 62.8 & 55.5 & 59.7 & \textbf{36.0} & 21.1 & \textbf{35.0} \\
\midrule
\textbf{\name (GPT-4o)} & 62.2 & 60.2 & 61.3 & 33.8 & 23.2 & 34.3 \\
\textbf{\name (o1-mini)} & \textbf{63.3} & \textbf{61.2} & \textbf{61.4} & 32.3 & \textbf{25.5} & 32.5 \\
\bottomrule
\end{tabular}
\label{cxr_result}
\end{table}

\subsubsection{Results and Analysis}

Table~\ref{cxr_result} presents the comparative performance of \name and baseline methods on the MIMIC-CXR chest X-ray classification task. 

The results demonstrate that \name configurations achieved strong performance across all evaluation metrics. 
The recent LLM-CXR approach demonstrates strong performance in Micro F1 score, slightly outperforming \name in this specific metric. However, \name shows substantially better performance in Macro AUC and Macro F1 scores, suggesting superior performance across all pathology classes, including less common conditions. Notably, \name dynamically orchestrates tool invocation based on reasoning needs, enabling adaptive analysis strategies. For example, when uncertainty arises in identifying cardiomegaly, \name can call additional agents for biometric retrieval or comparative analysis, enhancing robustness.
These results validate the effectiveness of \name' reasoning-driven tool utilization for chest-X-ray diagnosis, a task that requires both visual understanding and clinical reasoning. 

\subsection{Medical Visual Question Answering}
\subsubsection{Dataset and Task}

EHRXQA~\cite{bae2023ehrxqa} is a recently proposed multi-modal question answering benchmark that combines structured electronic health records (EHRs) from MIMIC-IV with chest X-ray images from MIMIC-CXR. The dataset includes three types of modality-based questions: Table-related, Image-related, and Image+Table-related, where the latter requires joint reasoning over both tabular data and medical images.

To ensure efficient evaluation while maintaining diversity, we randomly sample 300 QA instances from the EHRXQA test set, including 69 Image-related, 108 Table-related, and 123 Image+Table-related examples.

\subsubsection{Baselines}

We compare our method with two baselines:

\textbf{ChatGPT (Fixed):} \cite{bae2023ehrxqa} uses ChatGPT with a fixed prompt template and M3AE as the external VQA module.

\textbf{ChatGPT (BM25):} \cite{bae2023ehrxqa} retrieves few-shot examples from the training set using BM25 and uses them for in-context learning; M3AE remains the VQA backend.

It is important to note that these baselines rely heavily on a synthetic training dataset constructed from hand-crafted templates. Each question is paired with an executable program (SQL or NeuralSQL) and generated answer, filtered by execution validity. This design yields high supervision fidelity, but also requires access to database schema, image-label mappings, and fine-grained manual engineering.

In contrast, our method does not assume access to such structured supervision. We directly present each question to \name (GPT-4o), which autonomously reasons over the inputs and decides whether and how to invoke external tools to produce the final answer, without relying on any handcrafted training corpus or program annotations.
\subsubsection{Results and Analysis}

\begin{table}[t]
    \centering
    \caption{Performance comparison on EHRXQA (prediction execution accuracy). \name (o1-mini) achieves competitive results. It should be noted that the best baseline relies on two months of manual question-template–to-SQL (or NeuralSQL) conversion by four graduate students. In contrast, \name performs the entire pipeline automatically.}
    \label{tab:ehrxqa-results}
    \begin{tabular}{lccc}
        \toprule
        \textbf{Method} & \textbf{Image} & \textbf{Table} & \textbf{Image+Table} \\
        \midrule
        ChatGPT + M3AE (fixed)~\cite{bae2023ehrxqa} & 16.67 & 29.63 & 35.77 \\
        ChatGPT + M3AE (BM25)~\cite{bae2023ehrxqa} & \textbf{46.67} & \textbf{83.33} & \textbf{60.98} \\
        \textbf{\name (GPT-4o, best@5)} & 40.00 & 76.85 & 50.41 \\
        \textbf{\name (o1-mini, best@5)} & 41.67 & 78.70 & 54.47 \\
        \bottomrule
    \end{tabular}
\end{table}

As shown in Table~\ref{tab:ehrxqa-results}, our method achieves competitive performance across all modalities. In the table-related subset, \name demonstrates strong structured reasoning capability with an 78.70\% execution accuracy, although still behind the retrieval-based prompting baseline. For image-related questions, our model achieves a 41.67\% accuracy, indicating its basic visual perception and diagnostic ability. In the most challenging Image+Table setting, \name achieves 54.47\%, validating its capacity for cross-modal understanding.

\textbf{Advantages of \name.}
While \name does not yet surpass structured QA pipelines on this benchmark, it offers significant practical advantages in real-world settings. Unlike the benchmark baselines, which rely on a large-scale synthetic training corpus generated from handcrafted templates and executable programs, \name operates in a fully zero-shot or few-shot fashion and requires no manual efforts. Appendix \ref{sec:ehrxqabase} compares the two approaches in detail.

This distinction is critical in the medical domain, where manually constructing question-program-answer triples is infeasible at scale due to data privacy, annotation cost, and schema variability across institutions. By leveraging the reasoning capabilities, \name offers a generalizable and deployable solution that can flexibly respond to a wide range of natural questions about patients' medical records and imaging data, without requiring specialized retraining or specific engineering.

We argue that such agentic approaches better align with the long-term vision of general-purpose clinical AI assistants. They are not only more scalable but also more adaptable to unseen hospital systems, data schemas, and patient cohorts. 

\section{Conclusion}
We introduce \name, a goal-driven framework that represents a significant advancement toward highly customizable AI systems for medical decision support.

\name enables both autonomous and flexible customization of the entire reasoning workflow, allowing the system to dynamically coordinate diverse medical tools across different modalities and domains while adapting to specific clinical goals and institutional requirements.
In addition, it maintains transparent reasoning processes that record the entire decision-making pathway, allowing clinicians to audit and verify system recommendations. It also generates diverse reasoning pathways that provide healthcare professionals with multiple diagnostic perspectives beyond single conclusions.

Our evaluation across Alzheimer's disease assessment, chest X-ray diagnosis, and multimodal visual question answering demonstrates \name's competitive performance and high customizability.
By providing transparent, auditable reasoning processes and multiple diagnostic trajectories, \name enhances healthcare professionals' trust while supporting comprehensive clinical decision-making. This represents a paradigm shift toward interpretable and customizable frameworks that emulate clinical reasoning processes while enabling flexible adaptation to specific workflows and institutional requirements.

In the future, we aim to extend \name to broader clinical scenarios and incorporate additional medical modalities. Beyond healthcare, the framework's modular and customizable architecture holds promise for adaptation to other complex domains.

\bibliography{reference}

\bibliographystyle{unsrt}

\appendix
\section*{Appendix}

\section{Implementation Details}\label{sec:impl}

All experiments are conducted with GPT-4o and o1-mini as the core reasoning model.

\subsection{General-purpose Agents}
\paragraph{Web Search Agent.}
A lightweight wrapper around the Bing Search API returns the top–$k$ web hits (title, URL, date, snippet).  
The agent then fetches each page in parallel, parses HTML with \texttt{BeautifulSoup} (or PDFs with \texttt{pdfplumber}), and extracts a short context window around the snippet using a simple token-overlap heuristic.  
The reasoning process receives the resulting JSON block, rich enough for citation, yet small enough to keep the latency under two seconds per query.

\paragraph{Coding Agent.}
A self–contained tool lets the LLM draft short Python scripts, execute them in a sandbox, and stream the resulting stdout or error.  
The agent prompts the current backend model (GPT-4o, o1-mini, etc.) to emit \emph{code only}, strips any Markdown fences, saves the snippet to \texttt{temp.py}, and runs it via \texttt{subprocess}.  
This mechanism supplies the reasoning loop with on-the-fly calculations—statistics, small plots, and data reformatting without leaving the audit trail.

\paragraph{Text2SQL Agent.}
Our Text2SQL agent transforms natural language queries about patient information into structured SQL queries. The agent is provided with comprehensive schema information to ensure query compatibility. We employ a multi-step verification process wherein GPT-4o first generates the SQL query, then analyzes it for potential errors or inefficiencies before execution. The agent is equipped with specialized knowledge about medical database best practices, including handling of temporal queries (e.g., medication timelines, progression of symptoms) and appropriate joins across clinical tables (e.g., connecting patient demographics with laboratory results). 

\paragraph{RAG}
The Retrieval-Augmented Generation (RAG) system maintains a knowledge base of structured clinical guidelines from authoritative medical organizations. When queried, the system retrieves the most relevant guideline sections based on semantic similarity and recency.

\begin{figure}[h]
    \centering
    \includegraphics[width=0.95\textwidth]{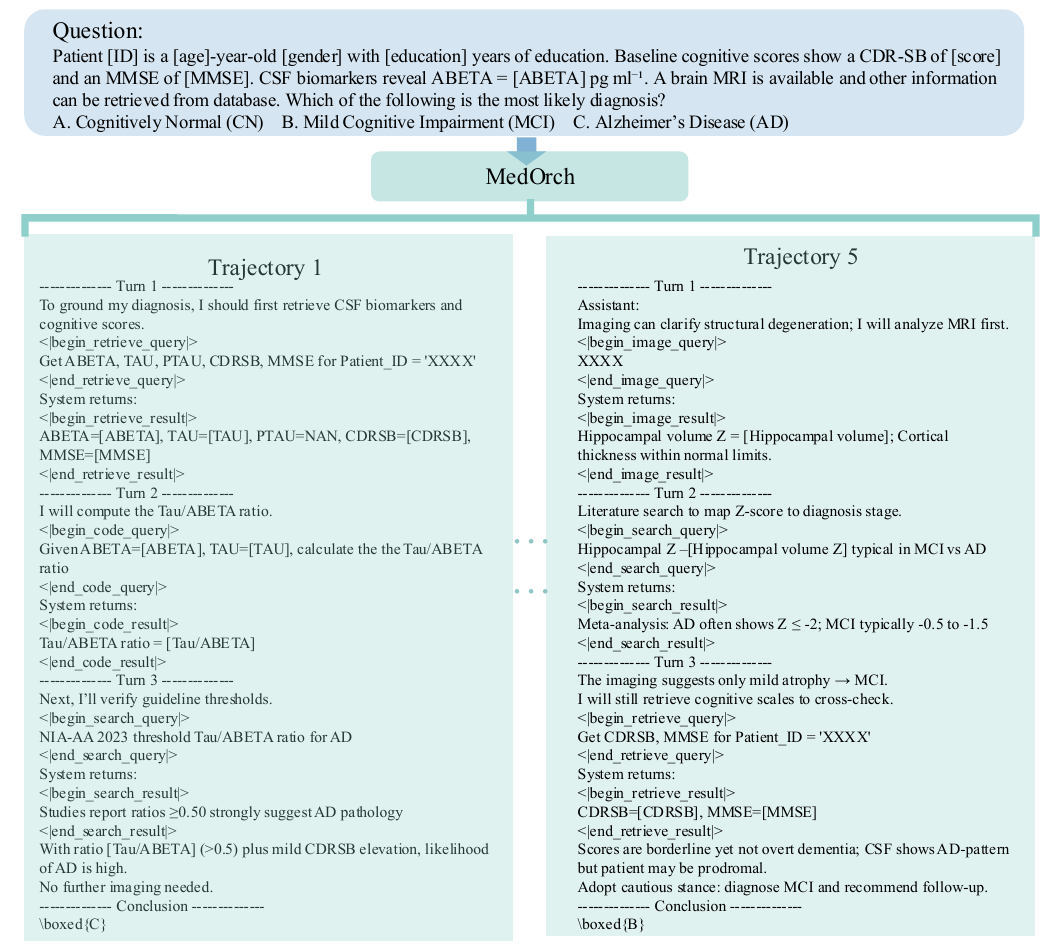}
    \caption{\name generates a set of transparent reasoning trajectories as references for clinicians}
    \label{fig:mult-tra}
\end{figure}

\subsection{Domain-specific Agents}
\textbf{Domain-specific agents for ADNI.}  
When working with the ADNI, we add two specialised modules.  
The \textbf{Medical Image Analysis Agent} wraps \texttt{FreeSurfer} for basic pre-processing and then applies a 3D ResNet-50 that we fine-tuned on baseline T1 scans.  A simple $80{:}15{:}5$ patient-level split yields a test accuracy of 84.34\%.  
Complementing this, the \textbf{Longitudinal Data Analysis Agent} first aligns each subject’s laboratory values, vital-sign records, cognitive scores, and imaging features to a common monthly timeline anchored at the baseline visit. Short gaps (\(<\) 3 months) are forward-filled, while longer gaps are linearly interpolated to preserve temporal trends. From these cleaned sequences the agent derives moving averages, first-order slopes, and simple rate-of-change ratios that summarise progression speed and direction. The resulting features are passed to the reasoning model, allowing it to reason over a patient’s longitudinal history rather than isolated snapshots.

\textbf{Domain-specific agents for \textsc{MIMIC-CXR} and \textsc{EHRXQA}.}
When analysing chest radiographs and multimodal clinical questions, we register two specialised modules.
The \textbf{Medical Imaging QA Agent} is composed of M3AE and GPT-4V. The \textbf{Clinical Knowledge-Graph Agent} processes each dialogue turn with UMLS to lift clinical concepts, and utilize GPT-4o to label the inter-concept relations. The resulting triples are streamed into a \texttt{nano\_graphrag} index that supports both local and global similarity search.

\section{Example of Multiple Reasoning Trajectories}\label{sec:mult_tra}

Figure~\ref{fig:mult-tra} shows \name autonomously produces a set of reasoning trajectories for the same question. Trajectory 1 begins by querying the electronic database for cerebrospinal-fluid (CSF) measurements and cognitive scores, computes the \textit{Tau/ABETA} ratio, and consults guideline literature before issuing an Alzheimer’s Disease (AD) diagnosis.  
Conversely, the Trajectory 5 launches an MRI volumetric analysis up front, interprets the hippocampal \emph{Z}\,-score with targeted literature search, and, after cross-checking cognitive scales, opts for a Mild Cognitive Impairment (MCI) assignment and recommends follow-up.  This makes three points explicit:  
(i) The same patient can yield distinct diagnostic conclusions when the reasoning strategy changes; 
(ii) \name’s modular orchestration of multimodal tools enables seamless strategy variation without touching the core engine; and  
(iii) Every intermediate calculation and evidence source is logged, affording full auditability to clinical reviewers.  
This substantiates our claim that multi-trajectory, transparent reasoning furnishes clinicians with diverse, evidence-linked perspectives in an inspectable framework.

\section{Comparision of \name and ChatGPT + M3AE (BM25)}\label{sec:ehrxqabase}

We present an example to show the advantage of \name over the ChatGPT + M3AE (BM25) method~\cite{bae2023ehrxqa}.

Here is a question template from EHRXQA:

\begin{tcolorbox}[
  colframe=blue!60,      
  colback=blue!5,       
  boxrule=0.8pt,         
  arc=2mm,               
  left=2mm, right=2mm,
  top=1mm, bottom=1mm,   
  enhanced,
  fontupper=\ttfamily\footnotesize, 
  breakable              
]
Given the [time\_filter\_exact1] study of patient [patient\_id] [time\_filter\_global1], is there
[attribute] in the [object]?
\end{tcolorbox}

The strongest baseline~\cite{bae2023ehrxqa} depends on \emph{hand-written} SQL/NeuralSQL templates.

\begin{enumerate}
  \item \textbf{Template-authoring stage.}  
        Four graduate annotators first examined the EHR dataset schema
        (202 tables spanning labs, imaging, demographics, etc.) and then spent
        about two months rewriting natural-language question
        templates into  
        (i) \textbf{SQL templates} for queries that require only relational
        data, and  
        (ii) \textbf{NeuralSQL templates} whose SQL clauses embed the special
        function \texttt{FUNC\_VQA(\dots)} to pass \texttt{study\_id} values
        into an M3AE image model for visual sub-questions.  
  \item \textbf{Inference stage.}  
        At run time, the system uses BM25 to retrieve the closest matching instances generated by templates, and asks ChatGPT to generate similar SQL or NeuralSQL to get answers for questions.
        Because the logical skeleton is fixed, any new question form or schema
        change requires writing or editing templates.
\end{enumerate}

Below is one such template pair for an EHRXQA question: 

\begin{tcolorbox}[
  colframe=blue!60,      
  colback=blue!20,      
  boxrule=0.8pt,         
  arc=2mm,              
  left=2mm, right=2mm,
  top=1mm, bottom=1mm,  
  enhanced,
  fontupper=\ttfamily\footnotesize,        
]

\textbf{Question Template:}\\
Given the \verb+[time_filter_exact1]+ study of patient \verb+{patient_id}+ \verb+[time_filter_global1]+, \textit{is there} \verb+${attribute}+ \textit{in the} \verb+${object}+\textit{?}

\vspace{0.8em}
\textbf{NeuralSQL Template:}
\begin{lstlisting}[language=SQL,basicstyle=\ttfamily\small,breaklines=true,frame=single]
SELECT FUNC_VQA("Is there ${attribute} in the ${object}?", T1.study_id)
FROM (
  SELECT tb_cxr.study_id
  FROM   tb_cxr
  WHERE  tb_cxr.subject_id = {patient_id}
    AND  tb_cxr.hadm_id IN (
           SELECT admissions.hadm_id
           FROM   admissions
           WHERE  admissions.subject_id = {patient_id}
         )
  time_filter_global1.event(admissions.admittime, admissions.dischtime)
  time_filter_global1_absolute(tb_cxr.studydatetime)
  time_filter_exact1(tb_cxr.studydatetime)
) AS T1;
\end{lstlisting}

\textbf{SQL Template:}
\begin{lstlisting}[language=SQL,basicstyle=\ttfamily\small,breaklines=true,frame=single]
SELECT MAX(T1.relation = 1)
FROM (
  SELECT *
  FROM   tb_cxr_plus
  WHERE  tb_cxr_plus.study_id IN (
           SELECT DISTINCT tb_cxr_plus.study_id
           FROM   tb_cxr_plus
           WHERE  tb_cxr_plus.subject_id = {patient_id}
             AND  tb_cxr_plus.hadm_id IN (
                    SELECT admissions.hadm_id
                    FROM   admissions
                    WHERE  admissions.subject_id = {patient_id}
                  )
           time_filter_global1.event(admissions.admittime, admissions.dischtime)
         )
  time_filter_global1_absolute(tb_cxr_plus.studydatetime)
  time_filter_exact1(tb_cxr_plus.studydatetime)
) AS T1
WHERE T1.object    = "${object}"
  AND T1.attribute = "${attribute}";
\end{lstlisting}

\end{tcolorbox}

In contrast, \name does not require any manual work and can be adapted to any other type of problem. Below is an example of the reasoning process of \name. We only give the question and necessary information to \name, and it reaches the final answer by reasoning and invoking agents or tools automatically.

\begin{tcolorbox}[
  colframe=blue!60,      
  colback=blue!20,       
  boxrule=0.8pt,         
  arc=2mm,              
  left=2mm, right=2mm,
  top=1mm, bottom=1mm,   
  enhanced,
  fontupper=\ttfamily\footnotesize,         
]
------------ Turn 1 ------------

To ground my answer, I need the study \verb+[patient_id]+'s chest X-ray taken within the specified window.

\verb+<|begin_retrieve_query|>+

Get \verb+study_id+ for \verb+Patient_ID = '[patient_id]'+ \\
within \verb+[time_filter_exact1 ± time_filter_global1]+. You should check the database schema before you generate the SQL.

\verb+<|end_retrieve_query|>+

System returns:

\verb+<|begin_retrieve_result|>+

\verb+study_id = [STUDY_ID]+ \\
\ldots

\verb+<|end_retrieve_result|>+

------------ Turn 2 ------------

With the study ID in hand, I will ask the medical image analysis agent whether the \verb+[object]+ shows \verb+[attribute]+.

\verb+<|begin_image_query|>+

\verb+[STUDY_ID]+

\verb+<|end_image_query|>+

System returns:

\verb+<|begin_image_result|>+

The image indicates \verb+[attribute]+.

\verb+<|end_image_result|>+

------------ Turn 3 ------------

To further confirm the answer, I should double-check with the medical-imaging QA agent.

\verb+<|begin_imageVQA_query|>+

\{``question'': ``Is there \verb+[attribute]+ in the \verb+[object]+?'',\\
\hspace*{1em}``image'': \verb+[STUDY_ID]+\}

\verb+<|end_imageVQA_query|>+

System returns:

\verb+<|begin_imageVQA_result|>+

answer = ``yes, \verb+[attribute]+ present in the upper right \verb+[object]+''

\verb+<|end_imageVQA_result|>+

------------ Conclusion ------------

Yes, \verb+[attribute]+ is present in \verb+[object]+.
\end{tcolorbox}

\end{document}